\pdfoutput=1
\documentclass{article}
\usepackage{spconf,amsmath,graphicx,hyperref}
\usepackage{cite}
\graphicspath{{figs/}}
\makeatletter %
\long\def\@makecaption#1#2{\vskip 6pt
 \setbox\@tempboxa\hbox{#1. #2}
 \ifdim \wd\@tempboxa >\hsize #1. #2\par \else \hbox
 to\hsize{\hfil\box\@tempboxa\hfil}\fi}
\makeatother

\title{Reference-Guided Turn-Taking Style Transfer \\for Full-Duplex Spoken Language Models}
\name{
Yuxuan Chen$^{1}$ \qquad
Yang Xiao$^{2}$ \qquad
Jack Sun$^{3}$ \qquad
Chak Sun$^{3}$ \qquad
Brick Li$^{3}$ \qquad
Sam$^{3,\dagger}$
\thanks{
$^{\dagger}$Corresponding author.
}
}

\address{
$^{1}$ The Chinese University of Hong Kong, Shenzhen\\
$^{2}$ The University of Melbourne \quad
$^{3}$ Lulula\\
yuxuanchen5@link.cuhk.edu.cn,\;
yang.xiao.1@student.unimelb.edu.au\\
jack@lulula.ai,\;
chak@lulula.ai,\;
brick@lulula.ai,\;
sam@lulula.ai
}
\begin{document}
\maketitle
\begin{abstract}
Full-duplex spoken language models decide at every frame whether to take
the floor, defining a conversational personality — yet each set of weights
encodes exactly one, which human speakers bracket rather than share.
Existing controls mostly act from outside; none identifies the quantity that
sets how much the model speaks.
We identify it: a single late-layer residual-stream direction, the
\textbf{initiative vector}, extracted from the model's own conversations
without training or annotation. The same direction reads the decision,
anticipating onsets up to 800\,ms ahead, and shifts it: scaled during
generation, it moves speech ratio from 6--12\% to 42--61\% (no steering:
21--30\%), consistently across conversations, weights, and corpora.
A short reference recording suffices to transfer a speaker's turn-taking
style to the model, and the transfer is reference-specific: swapping in
another speaker's reference worsens the match in 93--98\% of cases, whereas the
best uniform dose is indistinguishable from no steering.
Turn-taking style is not frozen into the weights but a direction that can
be read, shifted, and calibrated per person.
Audio examples are available at
\url{https://elpsykongloo.github.io/initiative-vector/}.
\end{abstract}


\section{Introduction}
\label{sec:intro}

\begin{figure}[t]
\centering
\includegraphics[width=\columnwidth]{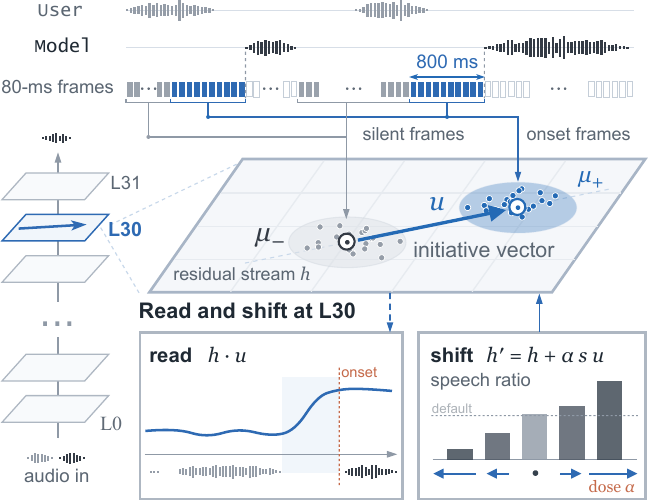}
\caption{The initiative vector: extracted from the model's own conversations,
read and written at L30. Top: among frames where the model has been silent,
the residual is averaged over frames followed by an onset within 800\,ms
($\boldsymbol{\mu}_{+}$) and over frames that stay silent
($\boldsymbol{\mu}_{-}$); their normalized difference is $\mathbf{u}$
(Eq.~\ref{eq:u}), with no training or annotation. Bottom: the projection
$\mathbf{h}\cdot\mathbf{u}$ rises before the model takes the floor, and adding
$\alpha s\mathbf{u}$ at every frame sets how much it speaks.}
\label{fig:overview}
\end{figure}

\begin{figure*}[t]
\centering
\includegraphics{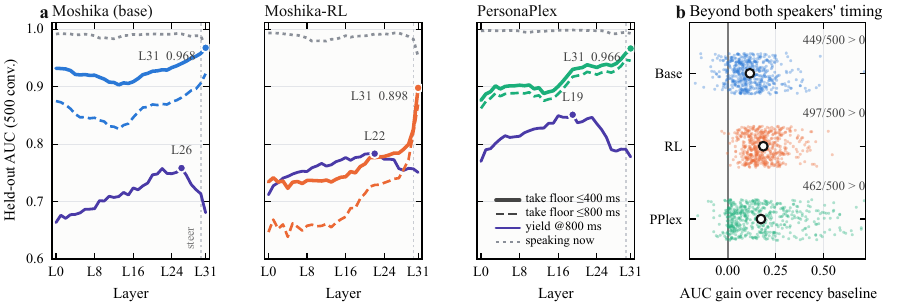}
\caption{The decision is linearly readable, deepest, and beyond both speakers'
timing. (a)~Held-out AUC of a class-mean direction refitted at each of the 32
residual positions (position $k$ is the output of block $k$); directions are
fitted on 100 conversations and evaluated on 500 held-out conversations per
weight. Taking the floor is strongest at the deepest positions on all three
weights (L31, 400\,ms: 0.968 / 0.898 / 0.966), whereas yielding peaks
mid-network and decays (L26 / L22 / L19): taking and yielding the floor are
not the same axis at the same depth. ``Speaking now'' is about 0.99 from L0,
i.e., present in the input. The vertical line marks L30, the injection
site. (b)~Per-conversation AUC gain of the fixed L30
direction over a timing baseline that sees both speakers
(Sec.~\ref{sec:setup}): $+0.115$ / $+0.185$ / $+0.172$, positive on 449 / 497
/ 462 of 500 conversations. Circles are means over conversations.}
\label{fig:read}
\end{figure*}

Spoken language models are converging toward full-duplex
interaction~\cite{defossez2024moshi,nguyen2022dgslm,veluri2024syncllm},
listening while speaking and deciding at every frame whether to take the
floor rather than waiting for an explicit
endpoint~\cite{skantze2021review,ekstedt2022vap}.
Collectively, this frame-level initiative defines a conversational
personality — eager or patient, interruptive or
deferential~\cite{chen2025fullduplexsurvey}
— and
different applications demand different ones: a tutoring agent should lead;
a companion should listen. Yet each set of weights encodes exactly one such
personality, fixed at training time~\cite{roy2026personaplex}, and no
default fits everybody: human speakers differ too much to share a
style~\cite{levinson2015timing,heldner2010pauses,cavalcanti2025dyadosyncrasy}
and fall on both sides of whichever one a checkpoint encodes. A uniform
correction therefore
helps one group and hurts the other — the best we find is indistinguishable
from leaving the model alone (Sec.~\ref{sec:results}). What each user needs
is personal~\cite{sutcliffe2024personasurvey}.

Most existing controls engineer turn-taking from the outside. Training-based
methods write a single style into the weights, from thousands of hours of
style-labeled data~\cite{zufle2026factor,pan2026lpstc} or from interactivity
rewards~\cite{tyagi2026steerduplex,ohashi2026kyutai,li2026duplexpo}: one
personality per checkpoint, bounded by the label set and prone to reward
hacking when timing itself is the reward. Instruction-based control avoids
retraining but is not reliably obeyed --- on
Instruct-FD~\cite{tang2026instructfd} the best system follows timing
instructions only 64.4\% of the time. Turn-taking
predictors~\cite{ekstedt2022vap,ekstedt2020turngpt,inoue2025promptvap} read the
relevant signal but cannot act on the model. Activation steering acts on the
model
itself~\cite{turner2023actadd,chen2025personavectors,chang2026perceptionvectors},
yet perception vectors set a binary listen/speak state at a fixed dose on one
checkpoint: a fix for interruption errors, not a variable to be set per person.

Whether a full-duplex spoken language model takes the floor is usually
treated as an emergent property of its next-token distribution, decodable
from internal states~\cite{riera2026sync} but never set by degree. A
single direction in the residual stream, the \textbf{initiative vector},
linearly encodes the model's propensity to
speak~\cite{park2024linearrep,arditi2024refusal}; scaling the residual along
that direction during generation controls that propensity. Because the
direction is computed from the model's own conversations, contrasting frames
that precede its speech onsets with frames where it stays
silent~\cite{panickssery2024caa}, each model yields its own vector under an
identical procedure (Fig.~\ref{fig:overview}).

\begin{figure}[t]
\centering
\includegraphics{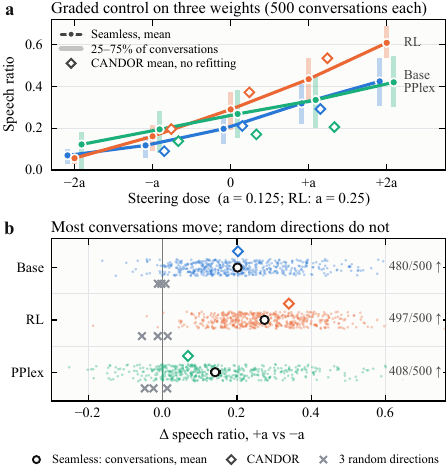}
\caption{Adding the direction changes behavior, gradedly and on a second
corpus. (a)~Speech ratio against steering dose, in units of each weight's
listen-to-speak swing $s$; 500 conversations per weight, and 500 CANDOR
segments with the same direction and scale. (b)~Paired per-conversation
difference between $+a$ and $-a$ under the same input and seed; random
directions are norm-matched and applied at the same displacement; black bars:
95\% intervals over conversations.}
\label{fig:shift}
\end{figure}

\begin{figure*}[t]
\centering
\includegraphics{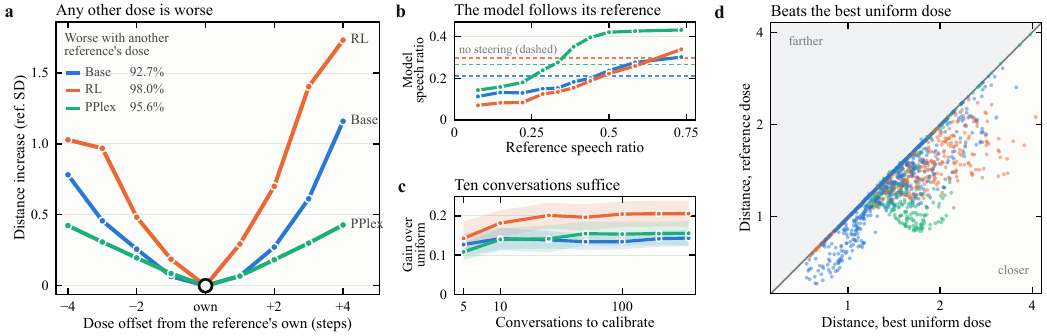}
\caption{One dose per speaker transfers that speaker's turn-taking style,
across three weights, 550 evaluation targets and 568 held-out references.
Distance (Eq.~\ref{eq:dist}) is in reference SD and averaged over three
seeds; each reference's dose is selected on 100 development targets only,
never on the targets used here. (a)~Increase in distance when a reference is
steered with a dose $k$ grid steps away from its own. (b)~Model speech ratio
after calibration, by decile of reference speech ratio. (c)~Gain over the
best uniform dose against the number of conversations used to select each
dose (300 targets; shading: 95\% interval). (d)~Distance under each held-out
reference's own dose against distance under the best uniform dose (568
references per weight).}
\label{fig:transfer}
\end{figure*}

The direction reads the decision to take the floor: its projection
anticipates the model's speech onsets well before they occur, above a
baseline built from both speakers' timing, and remains
predictive even when no recent text is present (Fig.~\ref{fig:read}b).
It also shifts the decision: as the steering dose increases, the model's
speech ratio rises monotonically, consistently
across conversations, weights, and corpora (Fig.~\ref{fig:shift}). Refitting
such contrasts at every layer locates the computation: whether the model is
currently speaking is decodable
from the first layer, evidence for yielding the floor peaks in earlier
layers and then fades, and evidence for taking the floor strengthens
toward the final layer (Fig.~\ref{fig:read}a). The model thus decides to take
the floor late in
the network, well after its speaking state is available and at a
different depth from its decision to yield. Calibrated against a short
reference recording, the direction transfers a speaker's turn-taking
style, and the transfer is reference-specific: neither a uniform dose nor
another speaker's reference matches as well (Fig.~\ref{fig:transfer}).

Our contributions are as follows:

\begin{enumerate}
\item \textbf{Initiative vectors.} A single residual-stream direction,
extracted without training or annotation, both anticipates the model's onsets
and sets how much it speaks — across weights and corpora.
\item \textbf{Taking the floor is a late-stage computation}, peaking in the
final layers, with the speaking state decodable in the first and yield
evidence peaking between them.
\item \textbf{Reference-guided style transfer.} A single short recording
suffices to transfer a speaker's turn-taking style, with no retraining;
swapping in another speaker's reference worsens the match in 93--98\% of
cases.
\end{enumerate}

\section{Method}
\label{sec:method}

\textbf{Extraction.} Let $\mathbf{h}_t$ be the residual stream at frame $t$
and position L30, the output of block 30 (zero-indexed) of the model's
32-block temporal transformer. From the model's own unsteered conversations we take the frames
at which it has been silent, and label a frame positive if the model starts
speaking within the next 800\,ms and negative if it stays silent. With class
means $\boldsymbol{\mu}_{+}$ and $\boldsymbol{\mu}_{-}$, the initiative
vector is
\begin{equation}
\mathbf{u}=(\boldsymbol{\mu}_{+}-\boldsymbol{\mu}_{-})\,/\,
\lVert\boldsymbol{\mu}_{+}-\boldsymbol{\mu}_{-}\rVert .
\label{eq:u}
\end{equation}
The labels are the model's own subsequent behavior, so extraction needs
neither training nor annotation.

\textbf{Reading and writing.} The projection $\langle\mathbf{h}_t,\mathbf{u}\rangle$
reads the model's propensity to take the floor. Steering writes along the same
direction at every frame during generation,
\begin{equation}
\mathbf{h}_t\leftarrow\mathbf{h}_t+\alpha\,s\,\mathbf{u},
\label{eq:steer}
\end{equation}
where $s$ is the listen-to-speak swing of the projection, the gap between its
mean while the model speaks and while only the partner speaks, so that the
dose $\alpha$ is comparable across weights. L30 is the deepest position whose
perturbation is still processed by a full transformer block and written into
that block's key--value cache; the shift therefore persists in the context
that later frames attend to.

\textbf{Depth profile.} To locate the decision, a class-mean contrast of the
same form is refitted at each of the 32 positions for four targets, each
scored by held-out AUC: taking the floor within 400 or 800\,ms of a frame at
which both speakers are silent, being silent 800\,ms after a frame of
sustained speech (yielding), and the current speaking state.

\textbf{Reference-guided transfer.} A speaker is summarized by three
descriptors of a five-minute reference recording: speech ratio, overlap
onsets per minute of partner speech, and response probability, the share of
partner turn ends (${\ge}400$\,ms of partner silence) followed within 3\,s by
the speaker's onset before the partner resumes, with turn ends during the
speaker's own speech kept in the denominator. With $\sigma_k$ the standard
deviation of descriptor $k$ over training references, descriptors
$\mathbf{b}$ and a reference $\mathbf{r}$ are compared by
\begin{equation}
d(\mathbf{b},\mathbf{r})=\Big(\frac{1}{3}\sum_{k=1}^{3}
\Big(\frac{b_k-r_k}{\sigma_k}\Big)^{2}\Big)^{1/2}.
\label{eq:dist}
\end{equation}
The speaker's dose minimizes the mean of $d(\mathbf{b}_t(\alpha),\mathbf{r})$
over development targets $t$, with $\mathbf{b}_t(\alpha)$ the steered model's
descriptors on $t$, and is then applied unchanged to new conversations. Since
$\mathbf{b}_t(\alpha)$ does not depend on the speaker, it is computed once
per weight, and a new speaker is calibrated from the recording alone, without
running the model.

\section{Experimental Setup}
\label{sec:setup}

\textbf{Models and conversations.} We study three full-duplex weights that
share the Moshi architecture~\cite{defossez2024moshi}: Moshika (base),
Moshika-RL, further aligned for interactivity~\cite{ohashi2026kyutai}, and
PersonaPlex~\cite{roy2026personaplex}, all operating on 80\,ms frames.
Each model hears one side of a dyadic Seamless Interaction
recording~\cite{seamless2025interaction} as its user stream (64\,s, after
3\,s of silence) and generates the other side. Directions and baselines are fitted on 100 conversations and evaluated on a
fixed set of 500 held-out conversations per weight. The same directions
and scale are applied, without refitting, to 500 CANDOR
segments~\cite{reece2023candor}. Doses are
$\{-2a,-a,0,a,2a\}$ with $a=0.125$ (Moshika, PersonaPlex) or $0.25$
(Moshika-RL).

\textbf{Readout.} AUC is computed per conversation and averaged. The baseline,
trained on the same 100 conversations, is a logistic regression on timing
alone: recent activity of both speakers, the time since each last spoke,
activity rates, and the time since the model's last text token.

\textbf{Transfer.} References are five-minute windows of Seamless
Interaction speakers from naturalistic and improvised conversations, with
descriptors extracted by Silero
VAD\footnote{\url{https://github.com/snakers4/silero-vad}}. Standard
deviations come from 120 training references, and the best uniform dose is
selected for them on 100 development targets; each individual dose uses the
same 100 targets alone, and transfer is evaluated on 550 targets for 568
held-out references, averaged over three seeds. Baselines are no steering and
the best uniform dose; the dose that is best in hindsight for each reference
serves as the upper bound. Intervals are 95\% two-way bootstrap intervals over
targets and references (5{,}000 resamples, stratified by conversation type,
doses fixed).

\section{Results}
\label{sec:results}

\textbf{The direction reads the decision ahead of time.} The fixed L30
projection separates frames followed by an onset within 800\,ms from continued
silence with AUC 0.818, 0.830 and 0.898 on Moshika, Moshika-RL and
PersonaPlex (Fig.~\ref{fig:read}b). The timing baseline, which sees both
speakers, reaches 0.703, 0.646 and 0.725; the direction adds $+0.115$,
$+0.185$ and $+0.172$ and is ahead on 449, 497 and 462 of 500 conversations.
The gain does not come from the text stream: on frames with no recent text
token the AUC stays at 0.806, 0.823 and 0.893, at most 0.012 below the full
set.

\textbf{Taking and yielding the floor peak at different depths.} Across the
32 positions (Fig.~\ref{fig:read}a), onset evidence is strongest at the
deepest ones on all three weights (0.968, 0.898 and 0.966 at L31 for
onsets within 400\,ms), whereas evidence for yielding the floor peaks
mid-network (L26, L22, L19) and then fades. Taking and yielding the floor are
thus separate axes at different depths. The speaking state is decodable at
about 0.99 from L0: it is present in the input rather than computed. The
model thus decides to take the floor late in the network.

\textbf{Adding the direction sets how much the model speaks.} Speech ratio
rises monotonically with the dose on all three weights, from 6--12\% at $-2a$
to 42--61\% at $+2a$, against 21--30\% without steering
(Fig.~\ref{fig:shift}a), and overlap onsets and response probability move in
the same direction. Under the same input and seed, moving from $-a$ to
$+a$ raises it by 20.1, 27.4 and 14.1 points, with increases on 480, 497 and
408 of 500 conversations (Fig.~\ref{fig:shift}b). The effect belongs to the
direction rather than to the size of the perturbation: three norm-matched
random directions at the same displacement change speech ratio by at most
5.6 points. The same direction and scale carry over to CANDOR without
refitting ($+20.3$, $+33.9$ and $+6.8$ points).

\textbf{One dose per speaker transfers that speaker's style.} On average,
the distance to a reference is smallest at its own dose on all three weights
(Fig.~\ref{fig:transfer}a). Swapping in another held-out reference's dose,
whenever it differs, worsens the match in 92.7, 98.0 and 95.6\% of cases,
and on average for 568, 568 and 566 of the 568 references. After calibration, the model's speech ratio follows its reference across
deciles (Spearman 0.69, 0.60, 0.78; Fig.~\ref{fig:transfer}b).

\textbf{The gain comes from personalization itself.} References fall on both
sides of every model's default (Fig.~\ref{fig:transfer}b), so no single
setting fits them: the best uniform dose is indistinguishable from no
steering on every weight, with all 95\% intervals including zero. The per-speaker dose improves on it by 0.129, 0.208 and 0.164 reference SD,
93.1, 99.5 and 98.1\% of the gain of the dose that is best in hindsight, and
it is strictly closer for 317/317, 257/257 and 270/289 of the
references whose dose differs from the uniform one (Fig.~\ref{fig:transfer}d).
Ten development conversations per weight already reach the plateau of this
gain (Fig.~\ref{fig:transfer}c). The dose a speaker needs is
also consistent across weights: the grid levels selected for the same
held-out reference correlate at Spearman $\rho=0.81$ (Moshika--RL), 0.77
(Moshika--PersonaPlex) and 0.70 (RL--PersonaPlex), so calibration captures a
property of the speaker rather than of one checkpoint.

\section{Conclusion}
\label{sec:conclusion}

A full-duplex model's turn-taking style need not be fixed at training time. A single late-layer direction, extracted from the model's own
conversations without training or annotation, anticipates when the model will
take the floor, sets how much it speaks across weights and corpora, and,
from a single short recording, transfers an individual speaker's style
without retraining. The benefit comes from personalization itself: turn-taking becomes a
setting that can be read, shifted, and calibrated per person.

\clearpage

\section{Compliance with Ethical Standards}
This study was conducted retrospectively using human subject data made
available for research by the Seamless
Interaction~\cite{seamless2025interaction} and CANDOR~\cite{reece2023candor}
corpora. Ethical approval was not required as confirmed by the licenses
attached with the data.

\bibliographystyle{IEEEbib}
{\small
\bibliography{strings,refs}

\begin{thebibliography}{10}

\bibitem{defossez2024moshi}
A.~D{\'e}fossez et~al.,
\newblock ``Moshi: a speech-text foundation model for real-time dialogue,''
\newblock {\em arXiv:2410.00037}, 2024.

\bibitem{nguyen2022dgslm}
T.~A. Nguyen et~al.,
\newblock ``Generative spoken dialogue language modeling,''
\newblock {\em Trans. Assoc. Comput. Linguistics}, vol. 11, pp. 250--266, 2023.

\bibitem{veluri2024syncllm}
B.~Veluri, B.~N. Peloquin, B.~Yu, H.~Gong, and S.~Gollakota,
\newblock ``Beyond turn-based interfaces: Synchronous {LLMs} as full-duplex
  dialogue agents,''
\newblock in {\em Proc. EMNLP}, 2024, pp. 21390--21402.

\bibitem{skantze2021review}
G.~Skantze,
\newblock ``Turn-taking in conversational systems and human-robot interaction:
  A review,''
\newblock {\em Comput. Speech Lang.}, vol. 67, pp. 101178, 2021.

\bibitem{ekstedt2022vap}
E.~Ekstedt and G.~Skantze,
\newblock ``Voice activity projection: Self-supervised learning of turn-taking
  events,''
\newblock in {\em Proc. Interspeech}, 2022, pp. 5190--5194.

\bibitem{chen2025fullduplexsurvey}
Y.~Chen and H.~Yu,
\newblock ``From turn-taking to synchronous dialogue: A survey of full-duplex
  spoken language models,''
\newblock {\em arXiv:2509.14515}, 2025.

\bibitem{roy2026personaplex}
R.~Roy et~al.,
\newblock ``{PersonaPlex}: Voice and role control for full duplex
  conversational speech models,''
\newblock {\em arXiv:2602.06053}, 2026.

\bibitem{levinson2015timing}
S.~C. Levinson and F.~Torreira,
\newblock ``Timing in turn-taking and its implications for processing models of
  language,''
\newblock {\em Front. Psychol.}, vol. 6, pp. 731, 2015.

\bibitem{heldner2010pauses}
M.~Heldner and J.~Edlund,
\newblock ``Pauses, gaps and overlaps in conversations,''
\newblock {\em J. Phonetics}, vol. 38, no. 4, pp. 555--568, 2010.

\bibitem{cavalcanti2025dyadosyncrasy}
J.~C. Cavalcanti and G.~Skantze,
\newblock ```{D}yadosyncrasy', idiosyncrasy and demographic factors in
  turn-taking,''
\newblock in {\em Proc. Interspeech}, 2025.

\bibitem{sutcliffe2024personasurvey}
R.~Sutcliffe,
\newblock ``A survey of personality, persona, and profile in conversational
  agents and chatbots,''
\newblock {\em arXiv:2401.00609}, 2024.

\bibitem{zufle2026factor}
M.~Z{\"u}fle, O.~Klejch, N.~Sanders, J.~Niehues, A.~Birch, and T.-K. Lam,
\newblock ``F-actor: Controllable conversational behaviour in full-duplex
  models,''
\newblock {\em arXiv:2601.11329}, 2026.

\bibitem{pan2026lpstc}
T.~Pan et~al.,
\newblock ``Enabling proactive spoken turns via a generalized style-aware
  full-duplex framework,''
\newblock {\em arXiv:2608.28630}, 2026.

\bibitem{tyagi2026steerduplex}
U.~Tyagi et~al.,
\newblock ``{SteerDuplex}: Steerable duplex speech dialogue models,''
\newblock {\em arXiv:2609.12623}, 2026.

\bibitem{ohashi2026kyutai}
A.~Ohashi, N.~Zeghidour, A.~D{\'e}fossez, and E.~Kharitonov,
\newblock ``Multi-faceted interactivity alignment in full-duplex speech
  models,''
\newblock {\em arXiv:2606.11167}, 2026.

\bibitem{li2026duplexpo}
Y.~Li et~al.,
\newblock ``Decoupling conversational dynamics in full-duplex spoken models
  through reinforcement learning,''
\newblock {\em arXiv:2607.07148}, 2026.

\bibitem{tang2026instructfd}
Y.~Tang et~al.,
\newblock ``{Instruct-FD}: Can your full-duplex speech system follow
  turn-taking instructions?,''
\newblock {\em arXiv:2607.20460}, 2026.

\bibitem{ekstedt2020turngpt}
E.~Ekstedt and G.~Skantze,
\newblock ``{TurnGPT}: a transformer-based language model for predicting
  turn-taking in spoken dialog,''
\newblock in {\em Findings ACL: EMNLP}, 2020, pp. 2981--2990.

\bibitem{inoue2025promptvap}
K.~Inoue et~al.,
\newblock ``Prompt-guided turn-taking prediction,''
\newblock in {\em Proc. SIGDIAL}, 2025.

\bibitem{turner2023actadd}
A.~M. Turner et~al.,
\newblock ``Steering language models with activation engineering,''
\newblock {\em arXiv:2308.10248}, 2023.

\bibitem{chen2025personavectors}
R.~Chen, A.~Arditi, H.~Sleight, O.~Evans, and J.~Lindsey,
\newblock ``Persona vectors: Monitoring and controlling character traits in
  language models,''
\newblock {\em arXiv:2507.21509}, 2025.

\bibitem{chang2026perceptionvectors}
C.-K. Chang, K.-W. Chang, A.~H. Liu, and J.~Glass,
\newblock ``Overcoming state inertia in full-duplex spoken language models via
  activation steering,''
\newblock {\em arXiv:2606.11386}, 2026.

\bibitem{riera2026sync}
P.~Riera, P.~Brusco, C.~Kuo, M.~Sancinetti, and S.~R.~K. Branavan,
\newblock ``Synchronization and turn-taking in full-duplex speech dialogue
  models,''
\newblock {\em arXiv:2605.20356}, 2026.

\bibitem{park2024linearrep}
K.~Park, Y.~J. Choe, and V.~Veitch,
\newblock ``The linear representation hypothesis and the geometry of large
  language models,''
\newblock in {\em Proc. ICML}, 2024, pp. 39643--39666.

\bibitem{arditi2024refusal}
A.~Arditi et~al.,
\newblock ``Refusal in language models is mediated by a single direction,''
\newblock in {\em Proc. NeurIPS}, 2024.

\bibitem{panickssery2024caa}
N.~Panickssery, N.~Gabrieli, J.~Schulz, M.~Tong, E.~Hubinger, and A.~M. Turner,
\newblock ``Steering {Llama} 2 via contrastive activation addition,''
\newblock in {\em Proc. ACL}, 2024, pp. 15504--15522.

\bibitem{seamless2025interaction}
V.~Agrawal et~al.,
\newblock ``Seamless interaction: Dyadic audiovisual motion modeling and
  large-scale dataset,''
\newblock {\em arXiv:2506.22554}, 2025.

\bibitem{reece2023candor}
A.~Reece et~al.,
\newblock ``The {CANDOR} corpus: Insights from a large multimodal dataset of
  naturalistic conversation,''
\newblock {\em Sci. Adv.}, vol. 9, no. 13, pp. eadf3197, 2023.

\end{thebibliography}
}

\end{document}